\documentclass[a4paper,12pt]{article}
\usepackage{multirow}
\usepackage{lmodern}
\usepackage{fullpage}
\usepackage{blindtext}
\usepackage{graphicx}
\usepackage{multicol}
\usepackage[utf8]{inputenc}
\usepackage{array}
\usepackage{float}
\usepackage{mathtools}  % brings in amsmath, also some improvements
\usepackage{amssymb} % brings in amsfonts, incl \square
\usepackage{amsmath}
\usepackage{layout}
\usepackage[algo2e, ruled, vlined]{algorithm2e} 
\usepackage{makecell}
\usepackage{algorithm}
\usepackage{algorithmic}
\usepackage{rotating}
\usepackage{tabularx}
\usepackage{tabularray}
\usepackage{geometry}
\usepackage{sectsty}
\usepackage{titlesec}
\usepackage{graphbox}
\usepackage[numbers]{natbib}
\usepackage{footmisc} % For additional footnote symbols
\usepackage{amsmath}  % For special symbols like \dagger
\usepackage[T1]{fontenc}
\titlespacing*{\section}      {0em}{.5em}{.3em}
\titlespacing*{\subsection}   {0em}{.50em}{.25em}
\titlespacing*{\subsubsection}{0em}{.50em}{.25em}
\titlespacing*{\paragraph}    {0em}{.5em}{.25em}
\usepackage[usenames,dvipsnames]{color}
\usepackage{indentfirst}
\usepackage[]{hyperref}

\titleformat*{\section}{\Large\bfseries}
\titleformat*{\subsection}{\large\bfseries}
\titleformat*{\paragraph}{\normalsize\bfseries}

\newcommand{\sur}[1]{\ensuremath{^{\textrm{#1}}}} % For using superscript with \sur{__text__}
\hypersetup{
  colorlinks,
  citecolor=WildStrawberry,
  linkcolor=WildStrawberry,
  urlcolor=WildStrawberry}

\title{\Huge{\textbf{CVCP-Fusion: On Implicit Depth Estimation for 3D Bounding Box Prediction}}}
\author{Pranav Gupta\footnotemark[2] , Rishabh Rengarajan\footnotemark[2] ,\\ Viren Bankapur, Vedansh Mannem,\\ Lakshit Ahuja\footnotemark[4] , Surya Vijay\footnotemark[4] , Kevin Wang\footnotemark[6]\\\normalsize{\textbf{Department of Computer Science, University of Texas at Dallas}}}
\date{June 2024}

\begin{document}
\maketitle
\renewcommand{\thefootnote}{\fnsymbol{footnote}}
\footnotetext[2]{These two authors contributed equally to this research.}
\footnotetext[4]{These two authors contributed equally to this research.}
\footnotetext[6]{Mentor of research.}

\begin{center} \Large{{\textbf{Abstract}}} \end{center}
\textit{Combining LiDAR and Camera-view data has become a common approach for 3D Object Detection. However, previous approaches combine the two input streams at a point-level, throwing away semantic information derived from camera features. In this paper we propose Cross-View Center Point-Fusion, a state-of-the-art model to perform 3D object detection by combining camera and LiDAR-derived features in the BEV space to preserve semantic density from the camera stream while incorporating spacial data from the LiDAR stream. Our  architecture utilizes aspects from previously established algorithms—Cross-View Transformers and CenterPoint—and runs their backbones in parallel, allowing efficient computation for real-time processing and application. In this paper we find that while an implicitly calculated depth-estimate may be sufficiently accurate  in a 2D map-view representation, explicitly calculated geometric and spacial information is needed for precise bounding box prediction in the 3D world-view space. Code to reproduce our results is available at  \url{https://github.com/safetylab24/FusionCVCP}.}

\begin{multicols}{2}

\sectionfont{\fontsize{15}{15}\selectfont}
\section{Introduction}
The 3D object-detection task has become commonplace when it comes to autonomous driving. As two fundamental modalities, LiDAR and camera sensors are capable of acquiring surroundings in different manners. LiDAR captures precise geometric and spatial information by providing an accurate 3D point cloud, while cameras contribute rich semantic details such as texture and color. However, previous approaches to multi-sensor fusion, particularly point-level fusion methods, fail to fully utilize the complementary strengths of these two modalities. Point-level fusion often leads to a loss of semantic information derived from camera data, resulting in sub-optimal performance in tasks that require precise 3D bounding box prediction and accurate object localization. Additionally, feature-level fusion may introduce geometric distortions when mapping LiDAR data onto a 2D image plane, which further limits robust detection, especially in tasks demanding high geometric accuracy.

In this paper, we propose Cross-View Center Point-Fusion (CVCP-Fusion) [\textit{figure \ref{fig:architecture}}], a state-of-the-art model designed to address these limitations by combining camera-derived semantic density with LiDAR's strong spatial data in the world-view (3D) representation. Unlike traditional anchor-based methods that struggle with rotational variance and require axis-aligned bounding boxes, our center-based approach simplifies detection by predicting object centers directly, making the system rotationally invariant and significantly reducing computational complexity.

Building on the established Cross-View Transformers\cite{CVT} and CenterPoint\cite{Center-Point}, our architecture runs their backbones in parallel, enabling real-time performance. The CenterPoint backbone is optimized for detecting object centers using LiDAR data, which are then refined in the second stage using keypoint detection to predict object properties such as size, orientation, and velocity. In contrast, the Cross-View Transformer enables the model to fuse multi-view camera features effectively into the BEV space, ensuring that the high-resolution semantic information is retained across modalities.

\textbf{Multi-Sensor Fusion. }One of the primary challenges in combining LiDAR and camera data is the discrepancy between perspective views from cameras and 3D views from LiDAR. Existing methods have typically focused on projecting LiDAR data into the camera view, introducing distortions that affect performance in geometric tasks such as 3D object detection and tracking. Our model addresses this by working in the shared BEV space, eliminating the need for perspective transformations that lead to information loss. By fusing multi-modal features within this unified space, CVCP-Fusion achieves higher detection accuracy while reducing computational overhead.

To efficiently utilize both input streams, the CVCP-Fusion architecture passes inputs through modal-specific encoders and combines extracted features in the 3D world-view space. CVCP-Fusion functions by extracting image features through an EfficientNet-B4 architecture and extracting LiDAR features through the Point Pillars Network\cite{PointPillars} 3D Encoder. This LiDAR encoding is simultaneously decoded into the 3D world-view  space and up-sampled using an RPN, which is finally concatenated to the decoded image embeddings outputted from the cross-view transformer.

\section{Related Works}

\paragraph{Depth Estimation. }
Many current implementations \cite{ BuildingRomeInADay, LonguetHiggins1981ACA, StructureFromMotion, PhotoTourism} utilize epipolar geometry and triangulation to explicitly compute camera extrinsics and depth. Despite being convenient, explicit depth calculations use multiple noisy inputs and require highly calibrated camera extrinsics data, often leading to inaccurate outputs for downstream calculations.

Monocular depth estimation aims to find depth information, rotation, and orientation to create a 3D representation of a view to allow for easier detection of features. Previously it has been shown that 2D estimations are significantly more robust than direct 3D estimates. Therefore, most algorithms \cite{Center-Point}, \cite{ROI-10D} first find features in 2D and regress to 3D using depth-estimation algorithms.  Center Point \cite{Center-Point} learns depth information directly while other algorithms such as ROI-10D \cite{ROI-10D} project to 3D to infer height. 

While monocular object detection does not need to consider inputs from multiple sources, it does require accurate depth estimation. Depth information is often difficult to predict and requires a probabilistic approach.  The Cross-View Transformers architecture addresses this by passing positional embeddings along with extracted features into transformer blocks, baking geometric reasoning into the weights of the transformer. This allows the transformer to convert a 2D multi-camera view input stream into a 3D world-view output while also incorporating information across the 6 different views.

\paragraph{Map-View Segmentation}\mbox{}\\
One common approach to Map-View Segmentation is using a Geometric Transformation on a planar image to convert it to a BEV \cite{Geotrans}, \cite{GeoTrans2}, \cite{GeoTransform}. However, more recent segmentation algorithms use a cross-view transformer architecture \cite{VPN}, \cite{CVT}, \cite{BEVsemseg}. This architecture achieves high accuracy with low computational costs making its implementation more feasible in real time. One such algorithm is VPN \cite{VPN}, which passes depth information and RGB data into a view-transformer module to create a semantic map of the scene. One drawback to this approach is that it doesn't consider camera calibration matrices, and relies on the model to learn them. Thus, we incorporate CVT \cite{CVT}, a state of the art algorithm that uses cross-view-attention to create a BEV using implicit geometric reasoning. Its primary advantage is that it uses positional embeddings to reduce inference time compared to explicit geometric reasoning.

\paragraph{3D BEV Object Detection}\mbox{}\\
3D Object Detection aims to create a BEV map-view segmentation of a scene which, in combination with height data, allows for 3D Object Detection. State of the art algorithms such as BEVDet \cite{BEVdet} use an image encoder-decoder architecture with a 3D object detection head to create bounding boxes. Similarly, M\sur{2}BEV \cite{M2BEV} uses an image encoder to learn image features and then projects to the BEV space, allowing for joint 3D object detection.

This research paper aims to achieve such results by combining two state-of-the-art algorithms, particularly CVT and CenterPoint, to create a 3D BEV Map-View segmentation and perform 3D object detection utilizing the embedded height data.

\section{Model Architecture}
\subsection{Cross-view Transformers}
Cross-view transformers aim to efficiently generate a map-view semantic representation of a vehicle's surroundings using an attention-based model. The architecture implicitly learns a mapping from individual camera views into a canonical map-view representation using a camera-aware cross-view attention mechanism. Each camera uses positional embeddings that depend on its intrinsic and extrinsic calibration. These embeddings allow a transformer to learn the mapping across different views without ever explicitly modeling it geometrically. The architecture consists of a convolutional image encoder for each view and cross-view transformer layers to infer a map-view semantic segmentation.

The height of any voxel can be approximated using the 2D birds-eye-view (BEV) label. Therefore, while other architectures focus on learning the geometric depth of any pixel in the 3D voxel-space relative to the position of the car, CVT learns depth as a proxy to the known height of the voxel. The cross-view transformer allows the network to learn any geometric transformation implicitly and directly from data. It learns an implicit estimate of depth through the camera-dependent map-view positional embedding by performing the downstream task as accurately as possible.

\subsection{Centerpoint}
CenterPoint is a state-of-the-art framework that predicts center-based 3D Object Detection and Tracking. It creates 3D bounding boxes by using voxel-space coordinates to represent, detect, and track 3D objects over a continuous-temporal state space. This allows the model to determine positional data along with movement data such as velocity. Centerpoint argues the main issue in accurate 3D bounding box detection lies in the inherent information loss when unprojecting from a 2D input to a 3D voxel space. To address this, for each detected center, Centerpoint regresses to all other object properties such as 3D size, orientation, and velocity from a point-feature at the center location. Furthermore, it uses a second stage to refine the object locations and extract point-features at the 3D centers of each face of the estimated objects 3D bounding box. This allows the model to recover the lost local geometric information due to striding and a limited receptive field, and brings a decent performance boost with minor cost.

\subsection{Cross-view Centerpoint}
CVT was trained on the nuScenes camera-only data, and it achieved state-of-the-art results on speed and accuracy. Since CVT learns a proxy for depth, we can directly feed its output, a BEV with height data embedded, into Center-Point as a proxy for the learned 3D embeddings. The two-stage CenterPoint block then computes the final 3D bounding box and its velocity. We use the VoxelNet network because prior results show that it outperforms the PointPillars 3D detection encoder.

\end{multicols}

\begin{figure}[t]
    \centering
    \includegraphics[width=16cm]{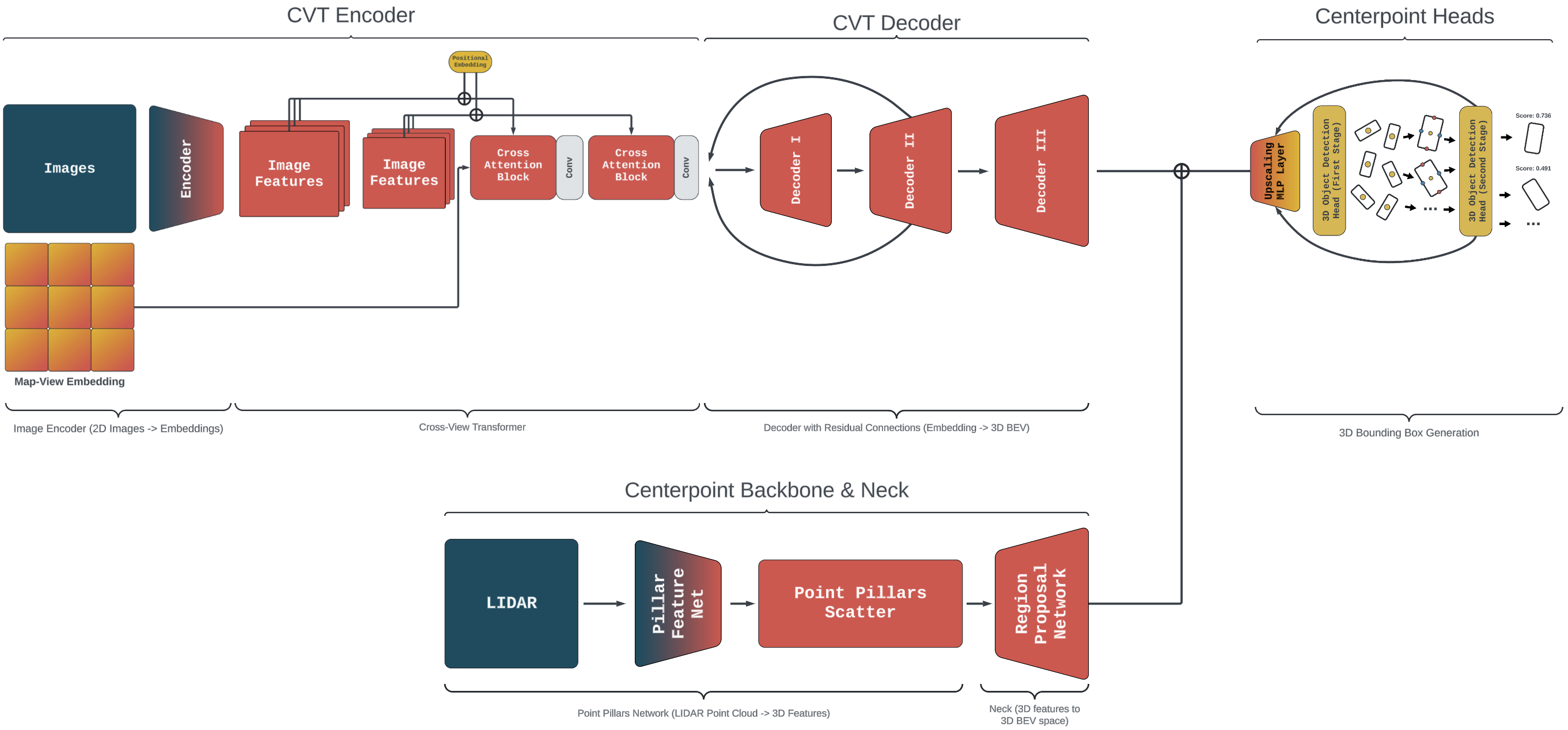}
    \caption{An overview of our proposed model architecture. For each image, we extract image features across multiple scales. Using known camera pose and intrinsics, we construct a camera-aware positional embedding. We learn a map-view positional embedding that aggregates information from all views through a series of cross attention layers. This is passed through a decoder which converts the learned embeddings into the 3D BEV space. In parallel, corresponding LiDAR data is passed through a Point-Pillars network and up-scaled using an MLP to the 3D BEV space. This is concatenated with the camera-derived BEV and convolved. Then, a 3D CNN-architecture detection-head finds object centers and regresses to full 3D bounding boxes using center features. This box prediction is used to extract point features at the 3D centers of each face of the estimated 3D bounding box, which are passed into an MLP to predict an IoU-guided confidence score and box regression refinement, allowing rotation of predicted 3D bounding boxes.}
    \label{fig:architecture}
\end{figure}

\begin{multicols}{2}

\section{Results}
We first present our 3D bounding box prediction results on the NuScenes test dataset for both the CVCP and LiDAR-Fusion models.

\textbf{Dataset} The nuScenes\cite{nuscenes} dataset contains 1000 diverse scenes ranging over a wide variety of times in the day, weather environments, and traffic conditions. Each scene contains 40 frames and lasts 20 seconds, for a total of 40,0000 data points in the dataset. Data encompasses the full 360º scene around the ego-vehicle captured through 6 camera views. Each camera view has calibrated intrinsics K and extrinsics (R,t) at every time-step.

We find that CVCP-Fusion doesn't perform well (mAP of 48.71 \ref{table}) compared to other modern implementations of 3D bounding box predictors. One major difference between these models is that they forego the error prone implicit depth calculations of CVT and instead use explicit depth calculations.

\bgroup
\def\arraystretch{1.5}
\begin{center}
\begin{tabular}{| c|c |} 
\hline
\large{Method} & \large{mAP} \\ 
\hline
PointPillars \cite{PointPillars} & 68.4\\ 
CVCNet \cite{CVCNet} & 82.7 \\ 
CBGS \cite{CBGS} & 81.1 \\ 
PointPainting \cite{PointPainting} & 77.9 \\ 
PMPNet \cite{PMPNet} & 79.7 \\ 
SSN \cite{SSN} & 80.7 \\ 
\hline
\textbf{CVCP} & >10.0 \\ 
\textbf{CVCP-Fusion} & 48.71 \\ 
\hline
\end{tabular}
\label{table}
\\
\textit{\\Table 1: State-of-the-art comparisons for 3D detection on nuScenes test set. We show the mAP and mAP for the car class.}
\end{center}
\egroup

\textbf{Implementation Details}
We tested two separate training methods when training the CVCP-Fusion model, one with a frozen CVT block and one training the entire model. For the former, we used two data-loaders to simultaneously pass in LiDAR and camera-view data. For the latter, we used two separate training sessions. The former to calculate decoder outputs from the CVT block [see top path in figure \ref{fig:architecture}], and the latter to calculate RPN outputs from the Centerpoint block [see bottom path in figure \ref{fig:architecture}], concatenate with the stored CVT tensors, and pass through the detection head. We saw similar performance in both training techniques and presented the latter approach's results as it provided a slightly higher mAP. 

We used a the ADAM optimizer, a base momentum of 0.85, a max momentum of 0.95, a learning rate of 0.001, and a batch size of 4 following Centerpoint's implementation details for their most successful session. The model took about about 24 GPU-hours when training on 3 GPUs for 15 epochs. After testing many possible values, we finally used a confidence threshold of 0.35 when evaluating the model to achieve our best results. Our hyperparameter settings can also be found in our config presets on \url{https://github.com/safetylab24/FusionCVCP}.

\end{multicols}
\begin{figure}[t]
{%
\setlength{\fboxsep}{0pt}%
\setlength{\fboxrule}{1pt}%
  \centering
  \fbox{\includegraphics[align=c,height=1.25in]{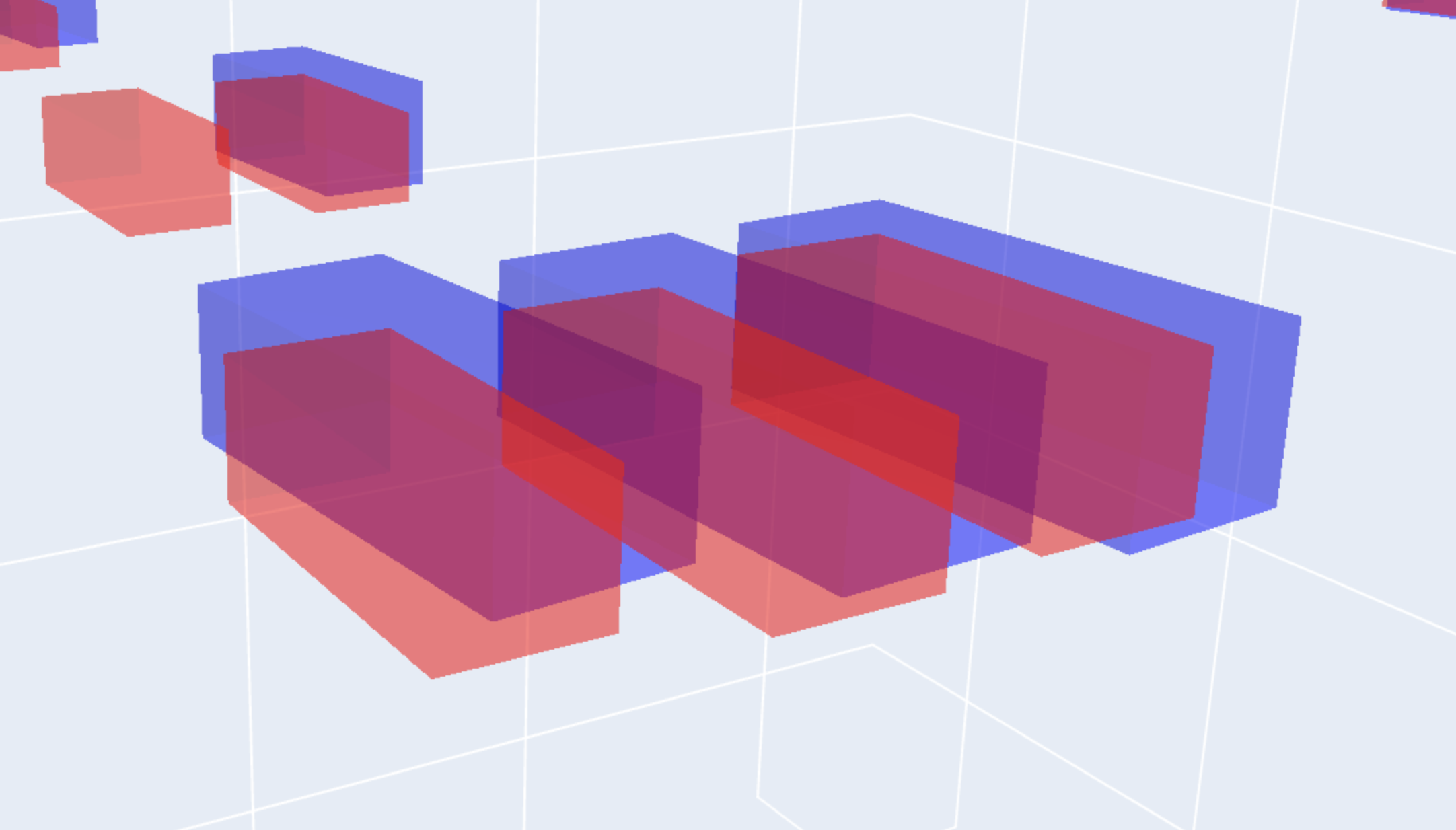}}
  \hspace*{.1in}
  \fbox{\includegraphics[align=c,height=1.25in]{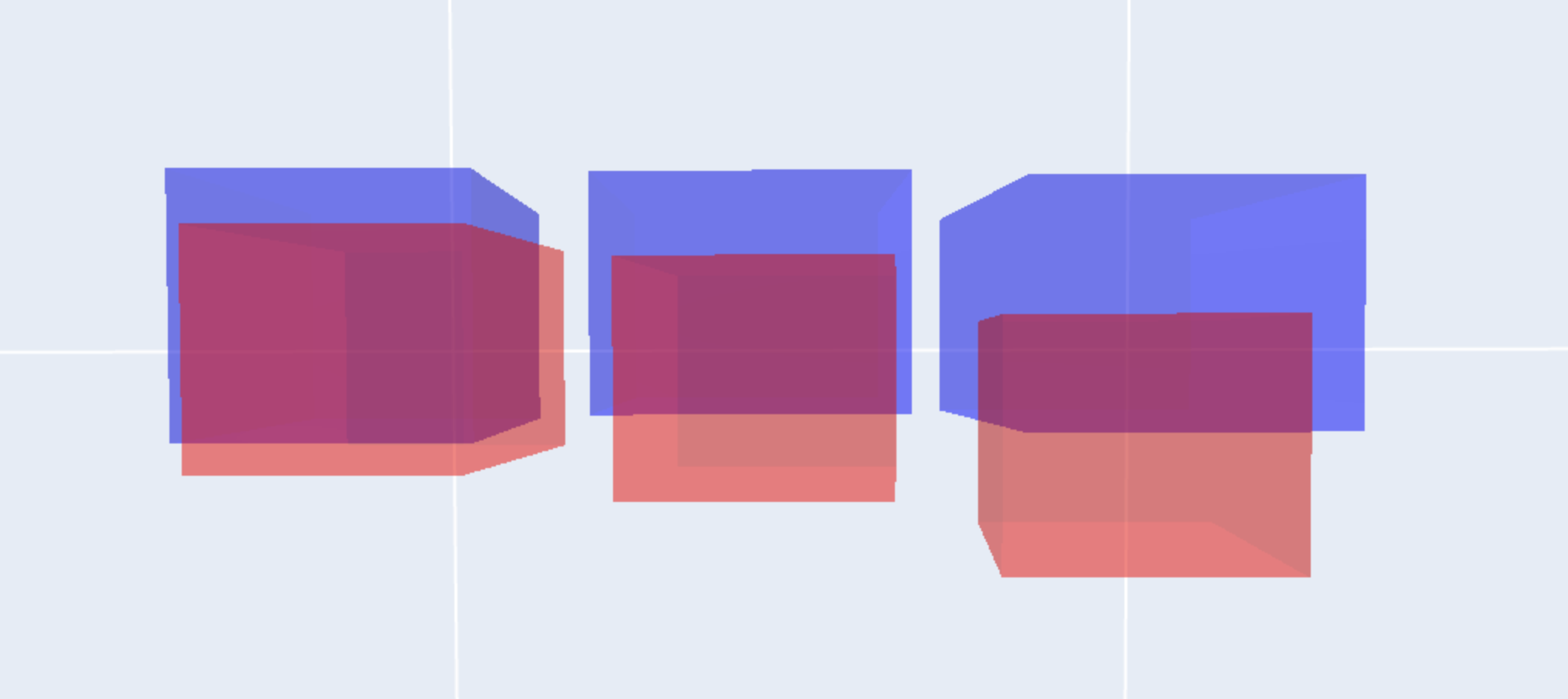}}
  \hspace*{.1in}
  \fbox{\includegraphics[align=c,height=1.25in]{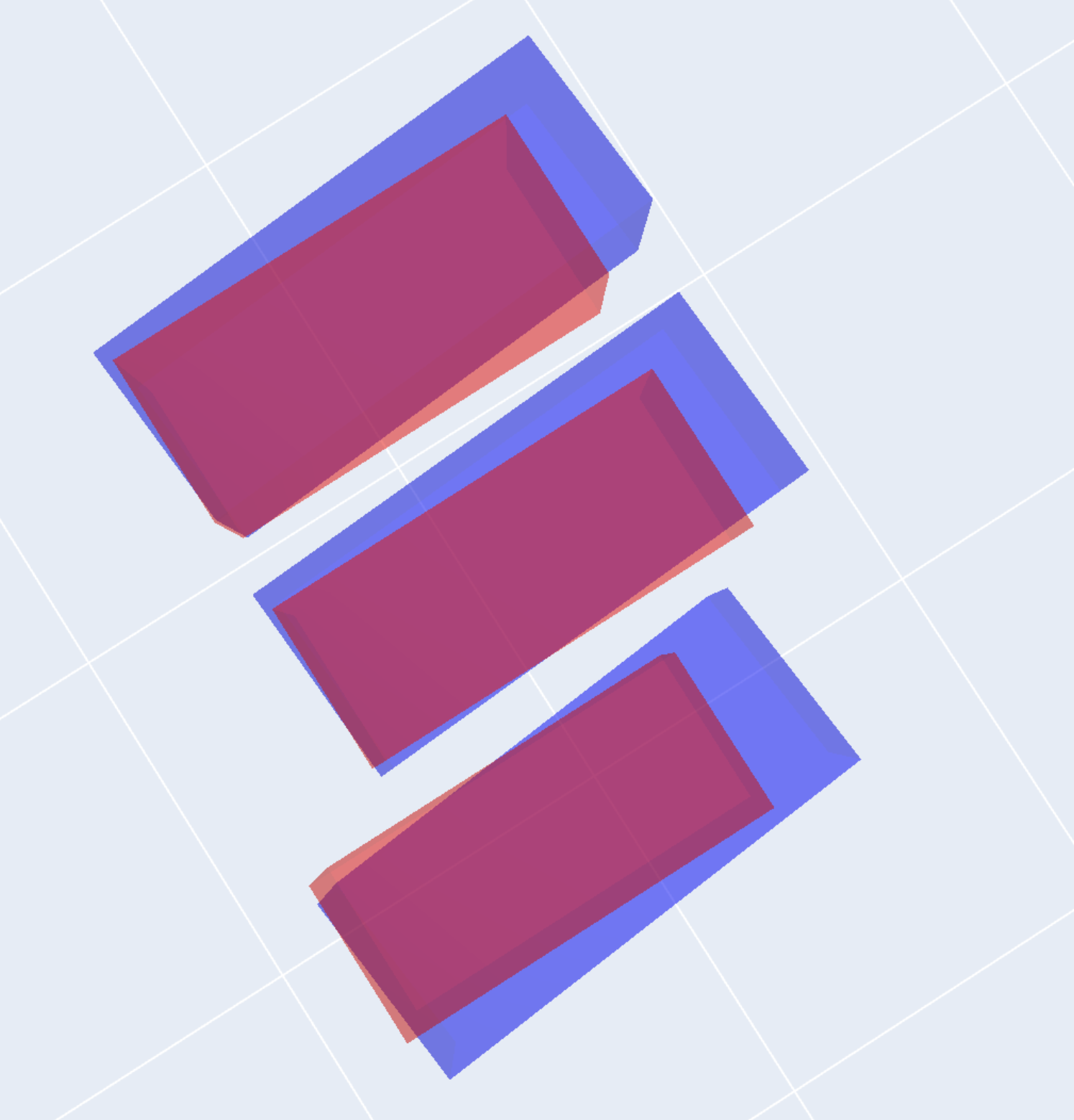}}
}%
\caption{Perspective, front, and top-down views, respectively. These three cars and their corresponding predictions sampled from the nuScenes dataset \textit{(ground truth in blue and predictions in red)} illustrate an example of CVCP-Fusion performing well at predicting rotation and bounding box coordinates in the x and y-dimensions, but failing to accurately predict in the the z-dimension.}
\label{predictionExample}
\end{figure}
\begin{multicols}{2}

\section{Analysis}
\subsection{Findings during Testing}
We find that bounding box predictions are inaccurate but consistently poor in a specific aspect: height. Going through predictions we found that CVCP-Fusion consistently accurately placed bounding boxes in the x and y directions but not on the z-axis. These three cars [figure \ref{predictionExample}] and their corresponding predictions sampled from the nuScenes dataset \textit{(ground truth in blue and predictions in red)} illustrate an example of this situation. While other factors such as false positives and imprecise rotation predictions also somewhat contributed to our fusion model's poor accuracy, height was consistently an issues the model struggled with the most. We found many more examples of this in almost every scene, with predictions being accurate in nearly every aspect except alignment in the z-axis.

\subsection{Speculation}
Attempting to extract 3-dimensional features from the output of the CVT decoder may be inherently flawed. The current transformer implementation accepts positional embeddings that were learned to map from a 2D projected view (i.e., the multiple camera-view inputs) to a 2D top-down view (i.e., the original BEV map-segmentation CVT was designed for). The current architecture may not allow accurate 3D features to be extracted from the transformer's output.

Furthermore, since depth is never explicitly calculated but is instead learned by the transformer, it isn't highly precise. The model's positional embeddings focus on ground-level relationships as they were designed to guide the model to output geometrically-informed outputs in the BEV space. This means that while height information \textit{can} be inferred through depth estimations, it will likely be highly inaccurate. I.e., while depth estimations are sufficient in a 2D output space, this indirect depth calculation may not be adequate to generating precise 3D predictions. 

Future models attempting to utilize cross-view transformers to generate 3D bounding boxes will likely need to encode vertical information into the positional embeddings along with X-Y information to guide the model to predict more accurately in the third dimension. Additionally, models could simply bypass this issue by using explicitly calculated depth through triangulation and stereo matching; CVT successfully attempts to implicitly learn depth, but this estimation isn't sufficiently precise in a 3D space. Finally, future models could also use explicit depth calculations and implicitly predict "adjustments" that are added onto the explicit calculation. This could combat compounding errors that often arise when using imprecise camera calibrations and noisy estimates while still maintaining a reliable \textit{general} depth estimate.

\subsection{Ablation Studies}
We tested our hypothesis that CVT couldn't perform accurately in the third dimension by removing the LiDAR block (i.e., moving to uni-modal input data, removing the concatenation, and adjusting layer sizes accordingly) of the CVCP-Fusion model to create CVCP. Training with both frozen and unfrozen weights yielded poor results as shown in [figure \ref{table}]. CVCP showed similar weaknesses in height predictions, but also failed to accurately predict the lateral (i.e., X-Y) position of objects. This suggests implicit depth calculations may be inherently unstable and may require large parameter sizes to perform well when being used in a larger dimension, which we leave to future work to test. This study further supports our speculation that CVT cannot accurately extract 3-dimensional features and struggles to provide precise depth calculations on larger scales without sufficient parameter count or embedding size.

\section{Conclusion}
CVCP-Fusion suggests that explicit depth calculations may be required for three-dimensional predictions. Implicit depth estimation using transformers is likely too inaccurate to be used for 3D bounding box prediction. Multi-modal fusion attempts and larger model sizes may remedy the inherent instabilities of relying on a model to bake accurate depth estimation into its weights.

\bibliographystyle{plain}
\bibliography{sample.bib}
\end{multicols}
\end{document}